\documentclass[journal]{IEEEtran}

\usepackage{cite}

\usepackage{graphicx}

\usepackage{amsmath}

\usepackage{algorithmic}

\usepackage{array}
\usepackage{booktabs}
\usepackage{colortbl}
\usepackage{multirow}

\usepackage[caption=false,font=footnotesize]{subfig}

\usepackage{dblfloatfix}

\usepackage{url}

\begin{document}

\title{(Pen-) Ultimate DNN Pruning}

\author{Marc~Riera,
        Jose-Maria~Arnau,
        and~Antonio~Gonz\'alez
\thanks{Department of Computer Architecture, Universitat Polit\`ecnica de Catalunya (UPC), Barcelona, Spain, contact e-mails: mriera@ac.upc.edu, jarnau@ac.upc.edu, antonio@ac.upc.edu}%
}

\maketitle

\begin{abstract}
DNN pruning reduces memory footprint and computational work of DNN-based solutions to improve performance and energy-efficiency. An effective pruning scheme should be able to systematically remove connections and/or neurons that are unnecessary or redundant, reducing the DNN size without any loss in accuracy. In this paper we show that prior pruning schemes require an extremely time-consuming iterative process that requires retraining the DNN many times to tune the pruning hyperparameters. We propose a DNN pruning scheme based on Principal Component Analysis and relative importance of each neuron's connection that automatically finds the optimized DNN in one shot without requiring hand-tuning of multiple parameters.
\end{abstract}


\section{Introduction}\label{s:Introduction}

DNN pruning has attracted the attention of the research community in recent years~\cite{Han2015, Han2016, Scalpel, SCNN}. Based on the observation that DNN models tend to be oversized and include a high degree of redundancy, pruning aims at reducing the model size by removing unimportant connections and/or neurons. The pruned model retains accuracy while requiring significantly less memory storage and computations, resulting in large performance improvements and energy savings.

Pruning requires the pruned model to be retrained; otherwise, the effectiveness of pruning is dramatically reduced. Finding the appropriate amount of pruning on each layer is a key factor that determines the efficiency of the scheme. If the pruning is too aggressive the DNN will not recover its accuracy after retraining, whereas if the pruning is too conservative, an opportunity to further optimize the DNN is lost. Previous DNN pruning schemes set the amount of pruning based on an expensive design space exploration or sensitivity analysis~\cite{Han2015}. We argue that these approaches are impractical for very deep neural networks, as they require retraining the DNN a large number of times, and each one may take several hours or days even on a high-end GPU for each retraining. To exacerbate the problem, hyperparameters such as learning rate or weight decay have to be manually tuned~\cite{Han2015, Scalpel}, further increasing the search space.

In this paper, we present a novel DNN pruning scheme that does not require such an expensive search to find the percentage of pruning for each layer, and it does not require to tune any hyperparameter. We refer to it as PCA plus Unimportant Connections (PCA+UC) pruning. Our scheme first applies node (i.e. neuron) pruning. We consider each layer as a system that produces ``information'' encoded by an N-dimensional array, where N is the number of neurons. The goal of pruning is to reduce the number of neurons to M (M~\textless~N) without loosing ``information''. The optimal pruning should find the minimum M for each level.

Principal Component Analysis (PCA) is a well-known statistical procedure that transforms a set of N-dimensional variables to a new coordinate system in which all coordinates are orthogonal and ordered from highest to lowest variance. Since variance can be considered as a good proxy of amount of ``information'', our scheme exploits PCA to determine the amount of neurons that can be safely removed. Although this is not the first work that proposes PCA for neuron pruning, we show that prior schemes~\cite{PruningPCA1, PruningPCA6} are ineffective for modern DNNs in Section~\ref{s:background}, and propose a different mechanism to effectively apply PCA-based node pruning in Section~\ref{s:pca_node_pruning}.

Once the percentage of pruning for a layer is set, our scheme has to select which neurons are removed. We have evaluated prior heuristics for node pruning~\cite{NodePruning,SimPruning} and found that, if retraining is applied, they achieve the same results than a blind node pruning that randomly selects the nodes to be removed from the model. Our experimental results show that the only relevant parameter is the amount of pruning, i.e. percentage of nodes to be pruned, and not which specific nodes are actually removed, since the topology of the DNN is not affected by that decision and the retraining will adjust the weights of the non-pruned nodes.

Once we determine the minimum number of neurons for each layer, there are still further opportunities to prune at the connection level. The first step, i.e. node pruning, results in layers that are not sparse since full neurons with all their connections are removed or kept. However, for a given non-pruned neuron there may be connections that are unimportant. After the PCA-based node pruning, our scheme measures the relative contribution of each connection with respect to the other connections of the same neuron. Those connections with a low contribution are removed and the final network is retrained. Unlike the node pruning step where the heuristic to select neurons is irrelevant, our results show that the heuristic used to choose the connections to be pruned has a non-negligible impact, achieving an additional 10\%-30\% of pruning over randomly choosing the connections. The overall scheme is non-iterative: it consists of only two steps and requires a single retraining after each of these steps. We show that this scheme produces results similar to or better than previously proposed iterative approaches that require an expensive (unfeasible for large networks) search. 

To summarize, this paper focuses on DNN pruning methodologies. We highlight the weaknesses of current pruning methods and solve them by proposing a more effective and practical scheme. The main contributions of this paper are the following:
\begin{itemize}
\item We analyze a selection of popular pruning methods and make two key observations. 
\begin{itemize}
\item First, for node pruning the heuristics used to decide what neurons to prune are irrelevant; the only important parameter is how much to prune, since the retraining adjusts the weights of the remaining neurons. On the other hand, the heuristic used for pruning connections is relevant as it may increase the amount of pruning by 10\%-30\% over a random selection.
\item Second, previous methods require an unrealistic number of parameters to be manually configured by trial and error for each DNN to be pruned.
\end{itemize}
\item We propose a novel, two-step pruning method that overcomes the above weaknesses. First, redundant nodes are removed by performing a PCA analysis of the outcome of each layer. Second, remaining unimportant connections are removed by taking into account their contribution relative to the rest of connections for a given neuron. PCA+UC provides a pruning of 72\% on average after evaluating it for multiple DNNs and is not iterative.
\end{itemize}

The rest of the paper is organized as follows. Section~\ref{s:background} reviews prior pruning methods. Section~\ref{s:pruning} presents an analysis of these pruning methods and a discussion about their effectiveness and practicality. Section~\ref{s:method} presents a new pruning method based on PCA and relative connection's importance. Section~\ref{s:methodology} describes the evaluation methodology. Section~\ref{s:results} discusses the experimental results. Section~\ref{s:related_work} reviews the related work. Finally, Section~\ref{s:conclusions} sums up the main conclusions of this work.

\section{Main DNN Pruning Schemes}\label{s:background}

A variety of different DNN pruning schemes have been recently proposed. Pruning reduces both the model size and the number of computations with the aim of reducing energy consumption and increasing performance. In general, pruning schemes require retraining of the pruned network to recover the original accuracy, and the resulting model becomes sparse which may incur in some overheads depending on the system, since working with sparse matrices/vectors is more costly than operating on dense arrays for many current systems.

Although pruning methods have achieved tremendous success for image classification (e.g. AlexNet~\cite{AlexNetV1,AlexNetV2}), there are very few studies about their effectiveness for other applications such as speech recognition. More importantly, all the schemes rely on heuristics with multiple parameters that require manual tuning, including the percentage of pruning for each layer. The next subsections provide more details on some of the most recent and popular pruning methods.

\subsection{Near Zero Weights Pruning}\label{s:near_zero_pruning}

Han et al.~\cite{Han2015} proposed a pruning method to remove the connections whose weight has an absolute value lower than a given threshold, which is computed using the following equation:
\begin{equation}
Threshold = std(W_{l})*qp
\end{equation}
where $std(W_{l})$ represents the standard deviation of all weights in layer \textit{l} and the quality parameter (\textit{qp}) determines the degree of pruning. The main idea of this heuristic is to remove the weights that are closer to zero. In the paper they report a 90\% pruning for AlexNet without accuracy loss. However, the quality parameter is different per layer and the paper does not present any methodology to set it up other than try and error. Note that exploring all possible combinations would be totally unfeasible for networks with many layers since each trial must be followed by a retraining, which is extremely expensive. Even for AlexNet, which has only 8 layers, the exploration of the design space is huge. For more recent DNNs such as ResNet~\cite{ResNet}, the winner of the 2015 Imagenet Large Scale Visual Recognition Challenge~\cite{ImageNet}, DenseNet~\cite{DenseNet} or SENet~\cite{SENet}, the winner for 2017, this would be impractical given that they have more than 100 layers.

We implemented this method using a global quality parameter for all the layers, to reduce the search space to just one parameter, and the degree of pruning achieved was more moderate, around 70\% in AlexNet. Finally, note that the heuristic they used does not work well if weights are not distributed around zero. In this case, this heuristic does not remove any connection, but there still may be connections that are unimportant compared to the others. For instance, if a neuron has ten input connections, nine of them have weights with a magnitude around 100 and the remaining one has a magnitude of around 10, this latter connection will likely be unimportant. However, this heuristic will not remove it since its weight is much larger than zero.

\subsection{Node Pruning}\label{s:avgw_pruning}

He et al.~\cite{NodePruning} proposed multiple metrics to identify which nodes are redundant for each layer $l$.

\begin{itemize}
\item Entropy:
\scriptsize
\begin{equation}
Score(i, l) = \frac{d^{l}_{i}(O)}{\left|O\right|}*\log_{2}(\frac{d^{l}_{i}(O)}{\left|O\right|}) + \frac{a^{l}_{i}(O)}{\left|O\right|}*\log_{2}(\frac{a^{l}_{i}(O)}{\left|O\right|})
\end{equation}
\normalsize
where $\left|O\right|$ is the total number of frames, and $a^{l}_{i}(O)$ and $d^{l}_{i}(O)$ are the number of frames which activate or deactivate node $i$.

\item Output Weights Norm (o-norm):
\scriptsize
\begin{equation}
Score(i, l, l') = \frac{1}{N^{l'}}\sum^{N^{l'}}_{j=1}\left|W^{l'}_{ij}\right|
\end{equation}
\normalsize
where $l'$ is the next layer for o-norm.

\item Input Weights Norm (i-norm):
\scriptsize
\begin{equation}
Score(i, l, l') = \frac{1}{N^{l'}}\sum^{N^{l'}}_{j=1}\left|W^{l}_{ji}\right|
\end{equation}
\normalsize
where $l'$ is the previous layer for i-norm.
\end{itemize}

The first metric, called Entropy, examines the activation distribution of each node. A node is considered activated if the output value is greater than a threshold of $0,5$. The idea of this metric is that if one node’s outputs are almost identical on all training data, these outputs do not generate variations to later layers and consequently they are not useful. The second and third metrics, called i/o-norm, determines the importance of a neuron based on the average of the weights of its incoming or outgoing connections. Nodes are sorted by their scores and those with lower scores are removed. The network is then retrained. 

All the metrics achieve similar results, around 60\% of pruning on TIMIT, a DNN for speech recognition. Note that one still has to decide how much to prune each layer, and they do not provide any heuristic to determine this other than trial and error. Furthermore, we show in Section~\ref{s:pruning} that a blind pruning that randomly selects the nodes to be removed achieves the same results as the aforementioned heuristics.

\subsection{Similarity Pruning}\label{s:sim_pruning}

Another way to detect redundancy, proposed by Srinivas et al.~\cite{SimPruning}, is to measure how similar the nodes are by computing the squared difference of the weights for each pair of nodes using the following equation:
\begin{equation}
Saliency(i, j, l) = \sum^{N}_{k =1}(\left\|W_{ik}-W_{jk}\right\|)^2
\end{equation}
The neurons with the lowest saliency are pruned. In this scheme, retraining is not applied but they achieve a rather moderate 35\% of pruning on AlexNet, which is low compared to other methods. This method is only applied to the fully-connected layers.

In short, this method avoids retraining but it achieves a low percentage of pruning and requires a huge space exploration to determine the particular threshold that should be used for each network layer.

\subsection{Scalpel}\label{s:scalpel_pruning}

Scalpel~\cite{Scalpel} is an iterative pruning method that determines which nodes to prune during training. To this end, a mask node is added after each original neuron to multiply its output by a parameter alpha that can be either 1 or 0. The method is divided into two steps, in the first step the mask layers are trained depending on a weight decay parameter that determines how much aggressive the pruning is. Then, the nodes for which the mask becomes zero after the training are removed and the network is retrained without the masks layers. This process is repeated multiple times, each time with an increased value of the weight decay, until a loss in accuracy is observed.

For the training phase when the masks are added, Scalpel uses two set of variables, called alphas and betas. Alphas represent the pruning mask that is applied to the output of the nodes $y_{i}$ during the forward evaluation of the network. That is: 
\begin{equation}
Y^{'}_{i} = \alpha_{i}*y_{i}
\end{equation}
where $Y^{'}_{i}$ are the final outputs passed to the next layer.

Alphas cannot be learned by the conventional Backpropagation method~\cite{rojas2013neural}, since the cost function is not a continuous function of alpha. To overcome this, Scalpel associates another parameter called beta to each alpha. These betas can take any Real value and are learned during training, as if they were the multiplicative coefficients applied to the outputs. 

Alphas are updated to 0 or 1 depending on the betas, a threshold and a epsilon offset as shown in equation~\ref{eq:scalpel}.
\begin{equation}
\label{eq:scalpel}
	\alpha_{i|k} = 
		\begin{cases} 
      1 & T+\epsilon \leq \beta_{i|k} \\
      \alpha_{i|k-1} & T \leq \beta_{i|k} < T+\epsilon \\
      0 & \beta_{i|k} < T
		\end{cases}
\end{equation}
Regularization is applied to the betas during training to penalize high values by using a weight decay parameter. The weight decay parameter also determines the amount of pruning since a high value will increase the penalization to the betas and the backpropagation algorithm will try to reduce them, and lower betas imply that more alphas are zero, which results in more pruning. Finally, multiple iterations of the algorithm are done by increasing the weight decay parameter on each iteration to increase the pruning, until a loss in accuracy is observed. The weight decay and the step that is used to increase it on each iteration have to be
manually set and are different for each DNN.

The main drawback of Scalpel is that it requires manual tuning of multiple parameters (threshold, epsilon, learning rate, etc.) for each particular DNN. Besides, its pruning effectiveness is not better than using previous methods, since its main target is pruning the DNN while avoiding sparsity. For instance, Scalpel achieves only 20\% of node pruning for AlexNet.

\subsection{PCA Pruning}\label{s:pca_pruning}

Levin et al.~\cite{PruningPCA1} proposed a pruning method to remove the nodes by using Principal Components Analysis (PCA). PCA can be used to reduce the number of nodes by computing the correlation matrix of the nodes activity. We can perform an eigendecomposition of the correlation matrix of the nodes activity to obtain the eigenvectors and eigenvalues. The eigenvalues can be used to rank the importance of a node of the new system.

The algorithm they propose to prune is divided into multiple steps. Starting from the first layer, the correlation matrix of the nodes activity is computed. The nodes activity is measured from multiple inputs of the training set and a pretrained network. The eigenvectors of the correlation matrix (i.e. Principal Components) are ranked by their corresponding eigenvalue and the effect of removing each node (i.e. eigenvector) is measured using the validation set. The nodes that do not increase the error are chosen to be removed. The weights of the layer are projected into the new subspace by multiplying the original weights ($W$) by the significant eigenvectors ($C_{l}$) as shown in equation~\ref{eq:old_pca1}. The procedure continues until all the layers are pruned. Note that this algorithm follows an iterative pruning since the validation and projection is performed after removing each node and stops when the accuracy of the network decreases. This method is only applied to the fully-connected layers and does not require any additional retraining.
\begin{equation}
\label{eq:old_pca1}
W -> W*C_{l}*C_{l}^{T}
\end{equation}
\begin{equation}
\label{eq:old_pca2}
W*I -> W*C_{l}*C_{l}^{T}*I
\end{equation}
Note that this scheme does not actually prune physical nodes or connections but reduces the number of parameters and computations depending on the amount of principal components that are removed. However, since the inputs also have to be projected using the significant eigenvectors as shown in equation~\ref{eq:old_pca2}, both the non-pruned eigenvectors and the projected weights have to be stored. Therefore, the pruning of eigenvectors has to be highly effective in order to actually reduce parameters and computations of the neural network.

The method was originally evaluated using an small feed-forward network of two layers with a time series dataset. We implemented this method for LeNet5 using the MNIST dataset and it achived only around 10\% of pruning with negligible accuracy loss. We have also tested this method on a modern Kaldi DNN~\cite{KaldiPaper} and the pruning achieved was less than 1\%. These low pruning percentages compared to previous methods suggest that pruning effectiveness is quite limited if the pruning method does not include retraining.

\section{Weaknesses of previous pruning schemes}\label{s:pruning}

\begin{table*}[t!]
\caption{DNNs employed for the pruning comparison. Kaldi is an MLP for acoustic scoring, AlexNet is a CNN for image classification and LeNet5 is a CNN for digit classification. The table only includes Fully-Connected (FC) and Convolutional (CONV) layers, as these layers take up the bulk of computations in DNNs. Other layers, such as ReLU or Pooling, are not shown for the sake of simplicity.}
\label{t:dnns}
\begin{center}
\begin{scriptsize}
\begin{sc}
\begin{tabular}{|cccc|cccc|cccc|}
\hline
\multicolumn{4}{|c}{\cellcolor[gray]{0.85}\textbf{Kaldi (18MB)}} & \multicolumn{4}{|c|}{\cellcolor[gray]{0.85}\textbf{AlexNet (200MB)}} & \multicolumn{4}{c|}{\cellcolor[gray]{0.85}\textbf{LeNet5 (12MB)}}\\
\multicolumn{4}{|c}{ Accuracy: 89.51\%} & \multicolumn{4}{|c|}{ Accuracy: 57.48\%} & \multicolumn{4}{c|}{ Accuracy: 99.34\%}\\
 Layer & Input Dim &\multicolumn{2}{c|}{ Output Dim} & Layer & In Dim & Out Dim & Kernel & Layer & In Dim & Out Dim & Kernel\\
 FC1 & 360 & \multicolumn{2}{c|}{ 360} & CONV1 & 3*224*224 & 64*55*55 & 11*11 & CONV1 & 1*28*28 & 32*28*28 & 5*5\\
 FC2 & 360 & \multicolumn{2}{c|}{ 2000} & CONV2 & 64*27*27 & 192*27*27 & 5*5 & CONV2 & 32*14*14 & 64*14*14 & 5*5\\
 FC3 & 400 & \multicolumn{2}{c|}{ 2000} & CONV3 & 192*13*13 & 384*13*13 & 3*3 & FC1 & 7*7*64 & 1024 & -\\
 FC4 & 400 & \multicolumn{2}{c|}{ 2000} & CONV4 & 384*13*13 & 384*13*13 & 3*3 & FC2 & 1024 & 10 & -\\
 FC5 & 400 & \multicolumn{2}{c|}{ 2000} & CONV5 & 384*13*13 & 256*13*13 & 3*3 & & & & \\
 FC6 & 400 & \multicolumn{2}{c|}{ 3482} & FC1 & 5*5*256 & 4096 & - & & & & \\
 &  &  &  &  FC2 & 4096 & 4096 & - & & & & \\
 &  &  &  &  FC3 & 4096 & 1000 & - & & & & \\
\hline
\end{tabular}
\end{sc}
\end{scriptsize}
\end{center}
\vskip -0.20in
\end{table*}

In this section, we highlight the main weaknesses of popular DNN pruning schemes. For quantitative evaluations, we use three different DNNs, whose parameters are shown in Table~\ref{t:dnns}. Kaldi is a Multi-Level Perceptron (MLP) for acoustic scoring, a key task of a speech recognition system. It takes as input a window of 9 frames of speech (current frame and the four previous and four next frames), where each frame is represented as an array of 40 features. Kaldi DNN generates the likelihoods of the 3482 senones, where a senone represents part of a phoneme. On the other hand, LeNet5 and AlexNet are popular Convolutional Neural Networks (CNNs). LeNet5 is a small CNN to classify written digits. Finally, AlexNet is a CNN for classifying color images into 1000 possible classes that range from different animals to various types of objects.

We first evaluate the effectiveness of the schemes previously proposed to select the connections and/or nodes that are removed from the model, and compared them with a blind pruning scheme that randomly selects the connections/nodes. We report the accuracy loss and the amount of pruning achieved for the networks shown in Table~\ref{t:dnns}. More specifically, we implemented the near-zero pruning (which applies to connections), and two of the node pruning methods: the similarity and the i-norm pruning (see Section~\ref{s:background}). Although similarity pruning does not use retraining, we include it in all the methods for a fair comparison. We report results for different overall pruning percentages starting from 10\% and increasing it by steps of 10\%. The parameters of each pruning scheme are manually adjusted to attain the target percentage of global pruning. For instance, for near-zero pruning, we tried different values of the \textit{quality parameter (qp)} until the target percentage of pruning was attained. Note that the percentage of pruning applied to each individual layer is determined by the particular heuristics used by each method and is not uniform across layers, i.e. some layers are pruned more aggressively than others.

\begin{figure}[t!]
\centering
\includegraphics[width=1\columnwidth]{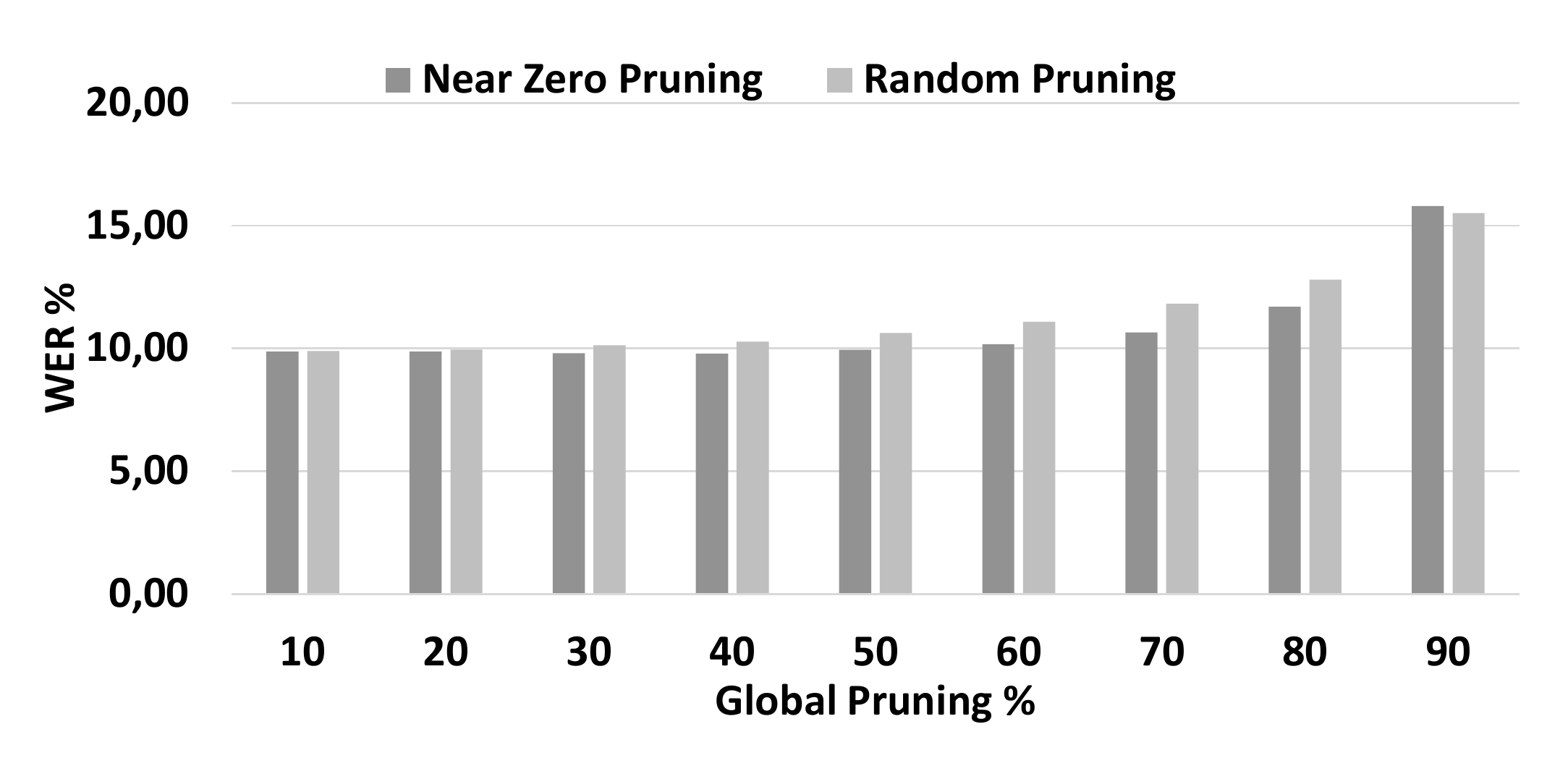}
\vskip -0.25in
\caption{Comparison between near-zero and random pruning of Kaldi DNN for different percentages of global pruning.}
\vskip -0.20in
\label{f:kaldi_linkp_cmp}
\end{figure}

Figure~\ref{f:kaldi_linkp_cmp} shows the comparison between near-zero and random pruning of connections in terms of Word Error Rate (WER is the main metric used in speech recognition; lower is better) for Kaldi DNN. We can observe that for 10-20\% pruning both methods achieve very similar accuracy. For 30-80\% pruning, near-zero is slightly better, and for 90\% pruning random is slightly more accurate. Random pruning achieves up to 50\% of pruning with negligible accuracy loss, whereas near-zero pruning achieves up to 70\%. Therefore, the near-zero scheme can achieve up to 20\% more pruning, but the random scheme still performs quite well. To further reinforce this conclusion, we performed multiple tests with the random scheme using different seeds to obtain different pruning patterns. For all the random experiments the accuracy obtained after retraining was almost the same, with smaller differences of less than 0.2\%.

\begin{figure}[t!]
\centering
\includegraphics[width=1\columnwidth]{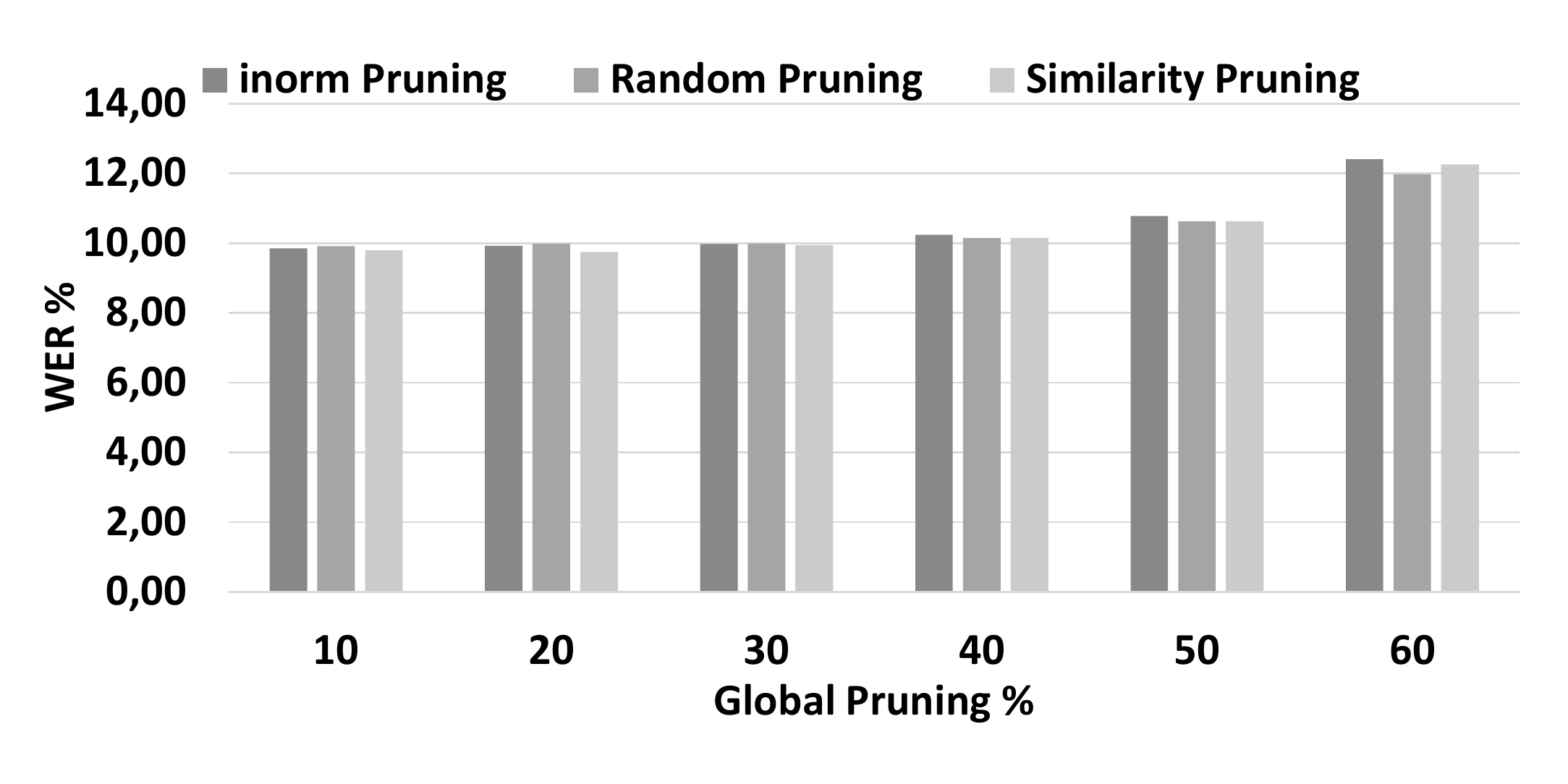}
\vskip -0.25in
\caption{Comparison between i-norm, similarity and random pruning of Kaldi for different percentages of global pruning.}
\vskip -0.20in
\label{f:kaldi_nodep_cmp}
\end{figure}

Figure~\ref{f:kaldi_nodep_cmp} shows the comparison between the different methods to prune nodes for the Kaldi DNN. We observe very minor differences in terms of accuracy among all the methods, so there is no clear winner. For node pruning, the last layer cannot be pruned since these neurons generate the output values that are used by the application. For instance, in Kaldi these are the probabilities used by the Viterbi beam search. Therefore, the maximum degree of pruning that can be achieved is around 60\% of the nodes. We can see that the differences in WER are less than 1\% in all the cases, random being slightly better when the global pruning is high (50\% and 60\%).

\begin{figure}[t!]
\centering
\includegraphics[width=1\columnwidth]{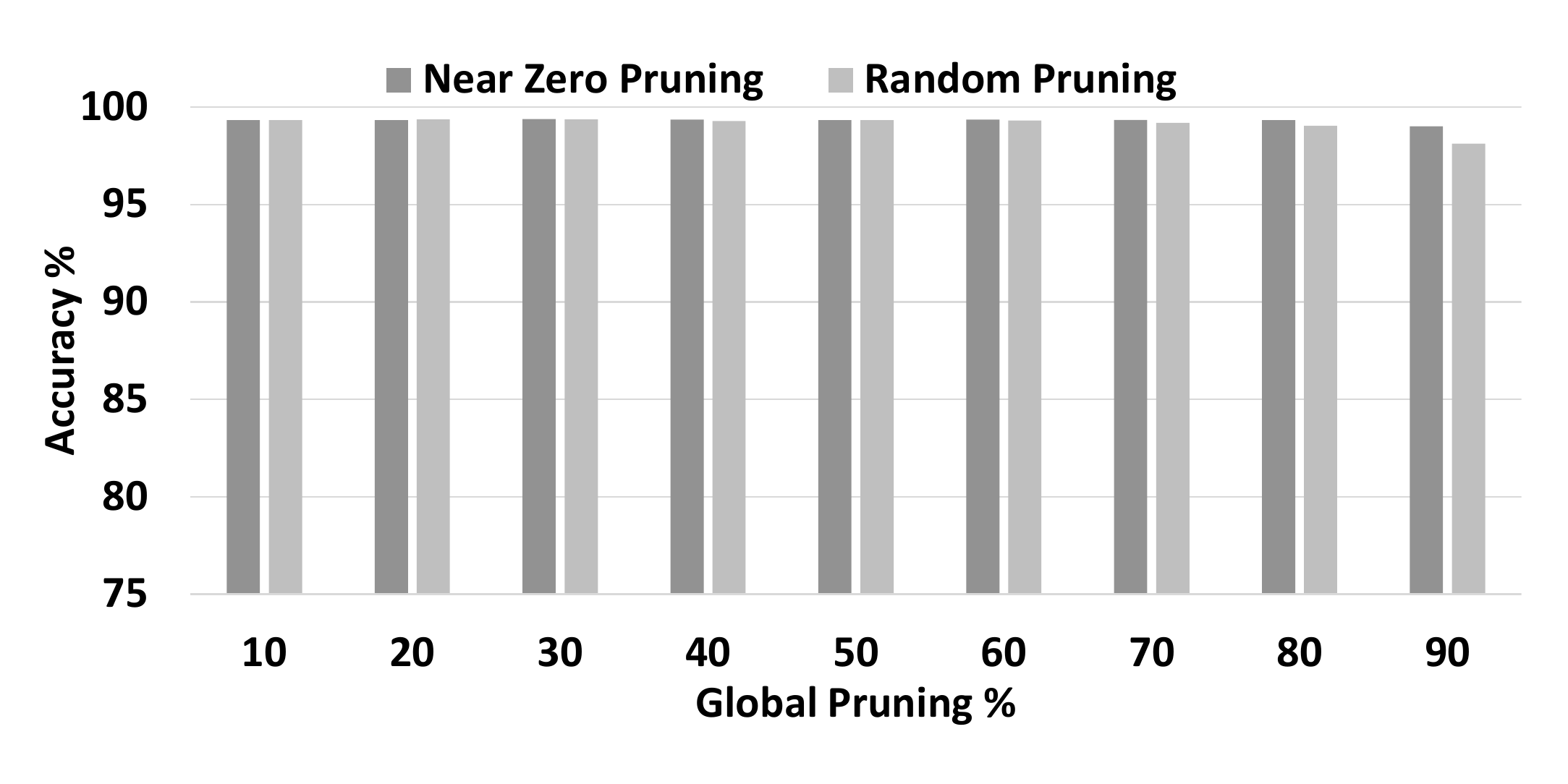}
\vskip -0.25in
\caption{Comparison between near-zero and random pruning of LeNet5 for different percentages of global pruning.}
\vskip -0.2in
\label{f:mnist_linkp_cmp}
\end{figure}

\begin{figure}[t!]
\centering
\includegraphics[width=1\columnwidth]{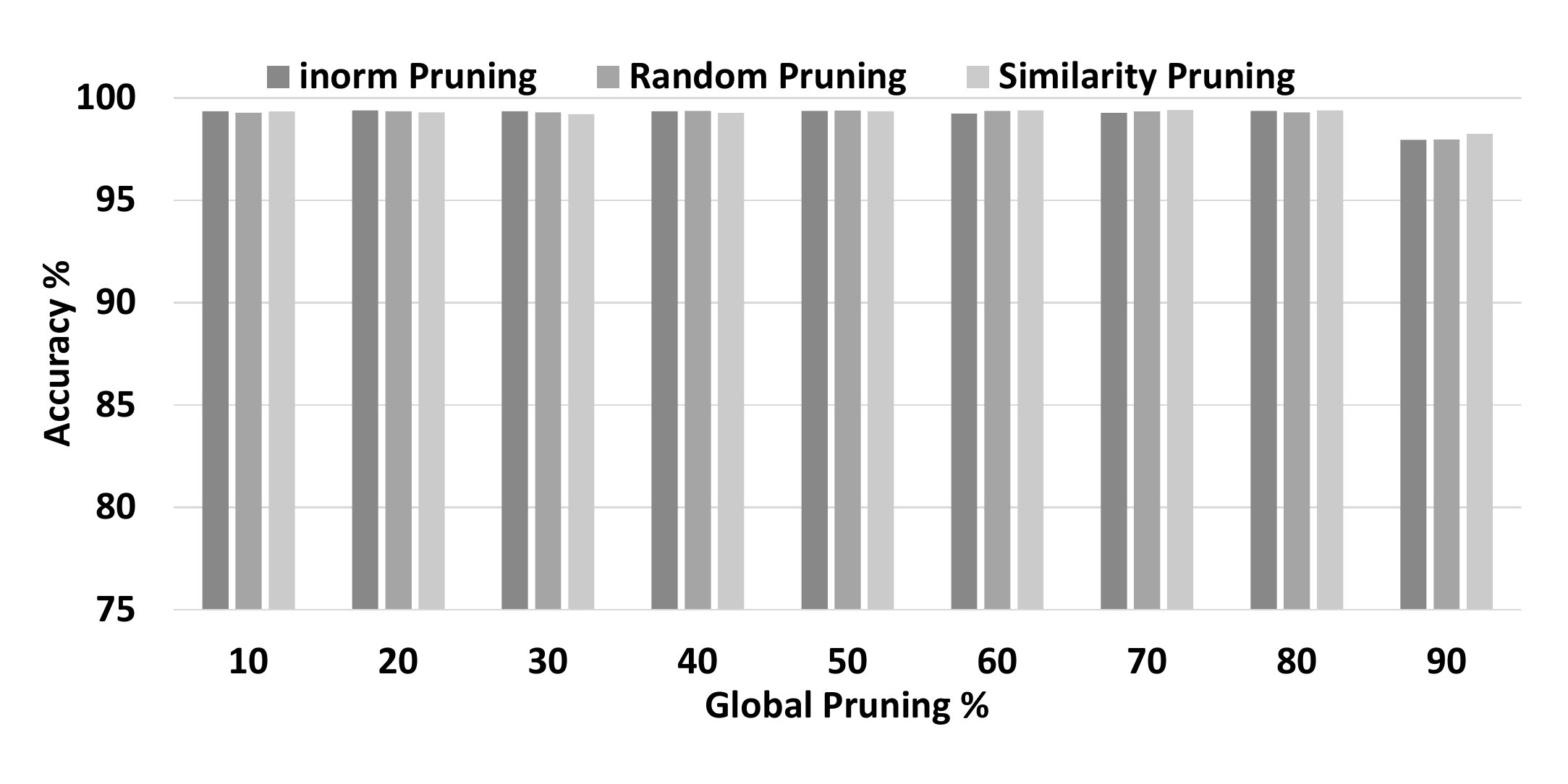}
\vskip -0.25in
\caption{Comparison between i-norm, similarity and random pruning of LeNet5 for different percentages of global pruning.}
\vskip -0.2in
\label{f:mnist_nodep_cmp}
\end{figure}

Results for LeNet5 are shown in Figure~\ref{f:mnist_linkp_cmp} for link pruning and Figure~\ref{f:mnist_nodep_cmp} for node pruning. In this case, LeNet5 has a high tolerance to errors and the accuracy is well maintained until pruning 90\% of the network for both types of pruning. At that point, the accuracy starts to decrease, being the random scheme slightly worse in the case of link pruning, but only by around 1\%, whereas for node pruning there are no significant differences between random, i-norm and similarity schemes for all percentages of pruning.

\begin{figure}[t!]
\centering
\includegraphics[width=1\columnwidth]{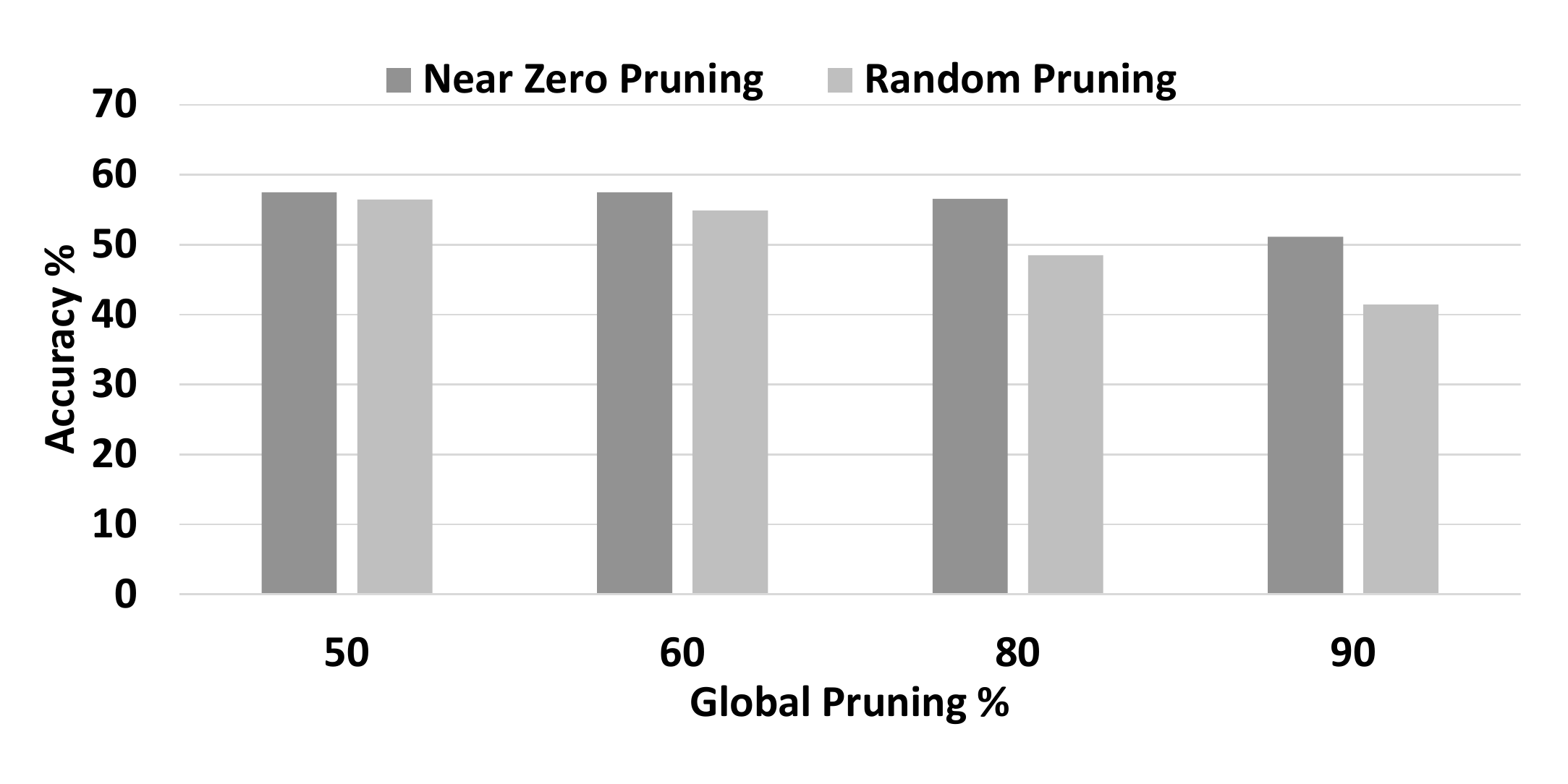}
\vskip -0.25in
\caption{Comparison between near-zero and random pruning of AlexNet for different percentages of global pruning.}
\vskip -0.2in
\label{f:alexnet_linkp_cmp}
\end{figure}

Figure~\ref{f:alexnet_linkp_cmp} shows the comparison between near-zero and random pruning of connections in terms of Top-1 accuracy for AlexNet. We can observe that up to 50\% of pruning the accuracy is largely recovered for both methods. Then, the random pruning starts to decrease the accuracy while the near-zero is able to maintain it until 80\%, and beyond that point it drops significantly.

Note that some studies in the literature report results for pruning that require the exploration of a huge design space, which may be impractical for some DNNs. For instance, the near-zero pruning scheme could obtain better results by applying a different quality parameter (\textit{qp}) for each network layer, as reported in the original paper~\cite{Han2015}. However, trying different values of \textit{qp} for each layer requires to explore an exponential number of configurations (exponential with the number of layers). Taking into account that for each configuration a retraining is needed, and retraining is extremely costly (can take several days in large networks), we claim that such strategies based on exploring an exponential number of configurations are impractical for many contemporary DNNs, which have hundreds of layers.

In summary, we observed that in terms of accuracy the random node pruning behaves almost equal to the rest of the methods. On the other hand, the random pruning of connections is somewhat less effective than the analyzed schemes. In other words, the heuristics to choose which neurons to prune are irrelevant while the heuristics to choose the connections may impact the final result. Besides, a main weakness of current pruning methods is their need of tuning one or multiple parameters through a trial and error process. This is extremely costly since each experiment requires retraining the network, and is impractical if the number of configurations to explore is too high (e.g. exponential with number of network layers).

\section{PCA+UC Pruning Method}\label{s:method}

Our proposed pruning method consists of two main steps. First, it performs a node pruning based on a PCA analysis of the data produced by each fully-connected layer. Next, some of the remaining connections are pruned based on their importance relative to the rest of incoming connections of the same neuron. The proposed scheme is not iterative and requires only two retraining operations, one after each of the two steps. We refer to it as PCA plus Unimportant Connections (PCA+UC) pruning.

\subsection{Node Pruning Through PCA}\label{s:pca_node_pruning}

The first step of the proposed approach is to prune redundant neurons in each layer through  a Principal Components Analysis (PCA)~\cite{PCA}. PCA is a well-known statistical method to summarize data, and is typically used to reduce the dimensionality of a dataset. PCA transforms a set of observations from different variables with high correlation into a set of principal components without linear correlation. Therefore, one of its main usages is to determine redundancy.

In our context, PCA is used to reduce the number of nodes of a layer. Each layer can be regarded as a system that for each evaluation generates an output value represented as a n-dimensional vector, where n is the number of neurons of this layer. PCA allows us to represent a set of n-dimensional values in a different coordinate system, without loosing any information, by applying a linear transformation:
\begin{gather}
\resizebox{0.80\columnwidth}{!}{$New_1 = Old_{1}*a_1+Old_{2}*b_1+\cdots+Old_{n}*z_1$} \nonumber\\
\resizebox{0.80\columnwidth}{!}{$New_2 = Old_{1}*a_2+Old_{2}*b_2+\cdots+Old_{n}*z_2$} \nonumber\\
\cdots \label{equation}\\
\resizebox{0.80\columnwidth}{!}{$New_n = Old_{1}*a_n+Old_{2}*b_n+\cdots+Old_{n}*z_n$} \nonumber
\end{gather}
Besides, in the new system the components are orthogonal (i.e., they do not contain any redundant information) and are ordered from higher to lower variance (i.e., from more to less information). If the original data presents high correlation among some of the n dimensions, in the transformed coordinate system, the last components will have very low variance. If we remove these low-variance components, we can represent the data in a lower-dimensional system with practically no loss of information.

In our case, we use PCA only to tell us how many neurons we need to preserve in each layer. The retraining process applied after pruning will adjust the weights so that the output of the pruned layer is equivalent to computing the original neurons and then applying the linear transformation dictated by the PCA. In other words, the original n-dimensional output is not needed, and the pruned network is expected to produce the same results as if the original n-dimensional output was computed and the PCA transformation was applied afterwards.

The steps to apply the PCA-based pruning are as follow. First, we generate a trace of the outputs of the nodes of each fully-connected layer (outputs are taken after the activation function). A subset of the training dataset can be used to generate the trace. In our experiments, we use around 1\% of the training set of each DNN to generate these traces. Then, we apply the PCA to the trace, which gives us the variance coefficients in the transformed coordinate system. Next, we compute how many of the lower-variance components can be removed while still keeping 95\% of the original variance. The number of remaining components is the number of neurons that we keep for this layer. Which nodes to keep and which are removed is irrelevant as demonstrated in the previous section, so they are randomly chosen. Once we determine the number of neurons in all layers, a single retraining of the pruned network is performed.

PCA cannot be directly applied to convolutional layers because a full feature map is considered as a single node which generates a volume of data. Since removing entire feature maps also have a high impact in accuracy we decided not to apply this step to convolutional layers. Note that previously proposed schemes for node pruning are also very inefficient for convolutional layers for the same reason.

\begin{figure}[t!]
\centering
\includegraphics[width=1\columnwidth]{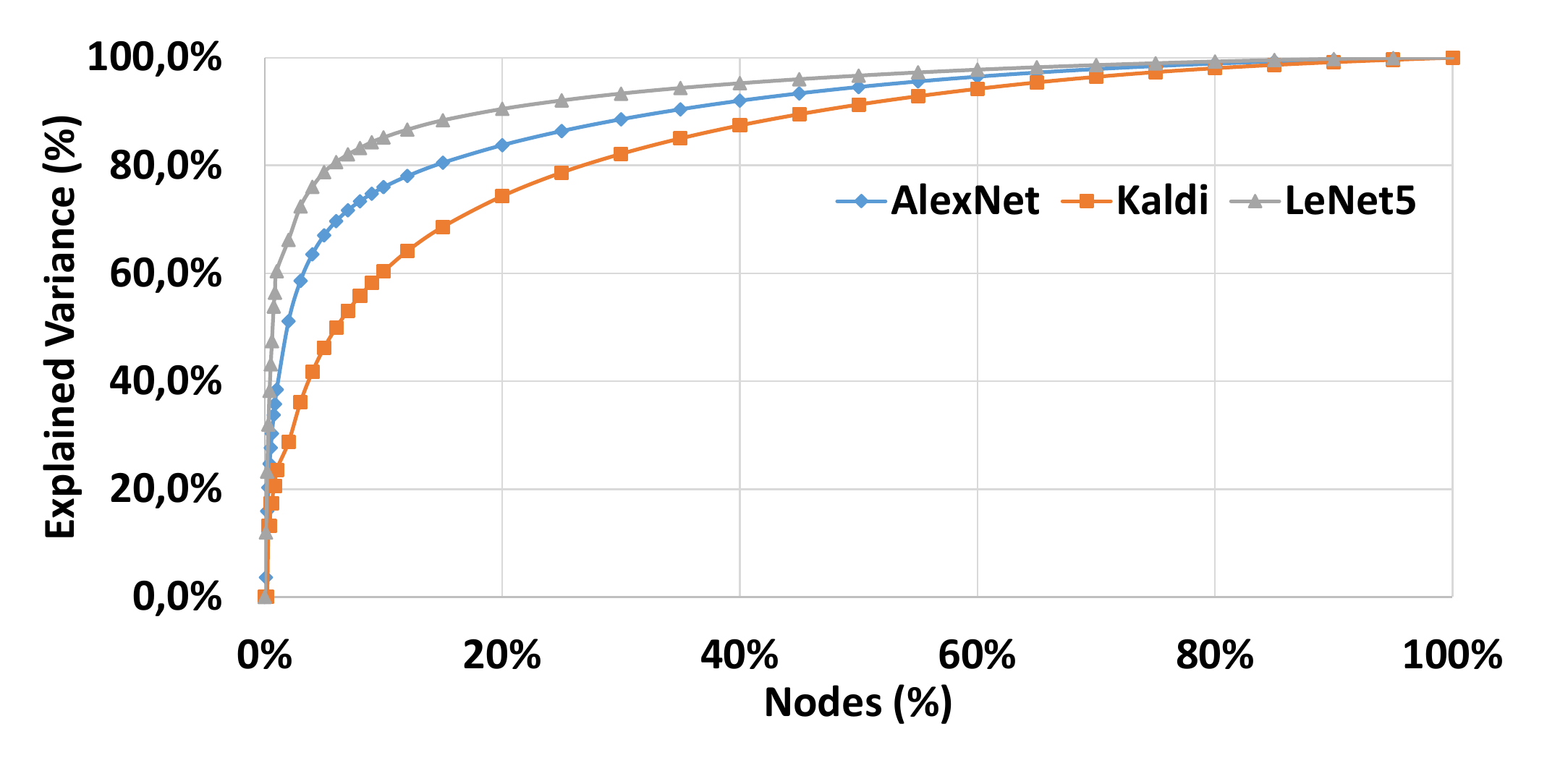}
\vskip -0.25in
\caption{Cumulative variance of a sample layer of AlexNet, Kaldi and LeNet5.}
\vskip -0.20in
\label{f:pca_var}
\end{figure}

Figure~\ref{f:pca_var} shows the cumulative variance of a sample fully-connected layer of AlexNet, Kaldi and LeNet5. For the sample layer of AlexNet, we can see that 50\% of the nodes keep 95\% of the original information, so 50\% can be pruned. For the sample layer of Kaldi, we can see that 60\% of the nodes keep 95\% of the original information, so 40\% can be pruned. Finally, for the sample layer of LeNet5, the benefits are much higher since we can remove 70\% of the nodes while still keeping 95\% of the information.

\subsection{Pruning Unimportant Connections}\label{s:pca_link_pruning}

The second step of our method consists of pruning unimportant connections after removing the redundant neurons. The rationale behind this approach is the following. The node pruning performed in the first step gives us a network with the minimal number of neurons in each fully-connected layer, while keeping the original accuracy. In the resulting topology there is no opportunity to remove further neurons; however, some of the connections may still have minor impact and can be removed to further reduce the size of the network. The obvious case are connections whose weight is zero. They clearly can be ignored without affecting the output. We could also remove all connections whose weight is close to zero, as the near-zero pruning~\cite{Han2015} does. However, the importance of a connection is not necessarily related to how close to zero its weight is. For instance, a weight of 0.1 would be unimportant if the rest of the weights for the same neuron are in the order of 1 or greater, but will be important if the rest of the weights are similar or smaller. In a similar manner, a connection with a weight very different to zero, say for instance 10, will be unimportant if the rest of the connections of the same neuron are in the order of 100.

We propose to measure the importance of a connection as a function of the absolute value of its weight and the average absolute value of all the incoming weights of the same neuron. In other words, a connection is considered unimportant if the magnitude of its weight is small compared with the other weights of the same node. This step is applied to all layers including convolutionals where a full feature map is considered as a single node.

A first idea to measure the importance of a connection would be to compute the ratio of the absolute value of its weight to the average absolute value of all weights of the same neuron. This works relatively well if the distributions of the weights are centered around zero for all neurons. Besides, this metric is insensitive to scaling. That is, if all weights are multiplied by a given constant, the pruning scheme would still affect the very same weights.

\begin{figure}[t!]
\centering
\includegraphics[width=1\columnwidth]{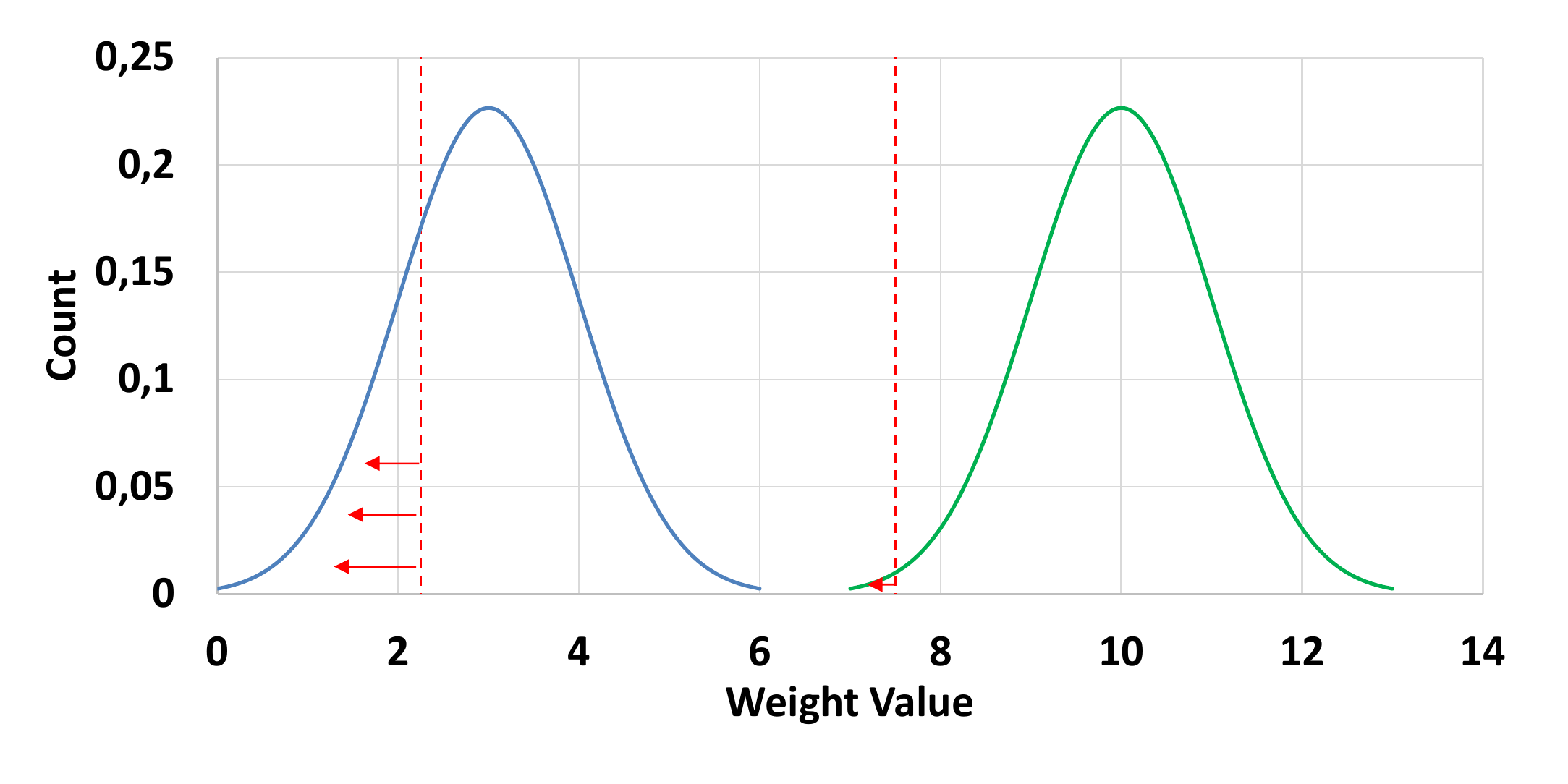}
\vskip -0.25in
\caption{Example of the weights distribution translation pruning problem.}
\vskip -0.20in
\label{f:invariant}
\end{figure}

However, this metric is not insensitive to displacements (translations). That is, if we have another neuron whose weights are about the same but with an added offset (i.e, each weight of the new neuron results from adding a constant to a different weight of the other neuron), this metric would result in a different pruning, in spite of the fact that the weight distribution of the two neurons have exactly the same shape, one being displaced by a constant with respect to the other. Figure~\ref{f:invariant} illustrates the translation problem with an example of two distributions of weights. Assuming a threshold of 75\% of the mean, the weights on the left side of the red line would be pruned in each case. We can see that the number of weights pruned for the distribution centered on three would be much higher than for the distribution centered on ten, although the only difference between them is a translation of seven.

We want a metric to measure the importance of a connection that is insensitive to translation and scaling of the weights. To this end, we first take the absolute value of all weights. Then, for each neuron, we substract the minimum absolute value of the weights of this node to the rest of the weights of that node. Finally, for each node we compute the mean of the resulting values and remove the connections whose value is smaller than 75\% of the mean.

An observation to make is that after applying the pruning of connections the resulting network model will be sparse. Executing a sparse model is normally less efficient than a dense model so the pruning ratios achieved by this step must be significant to compensate for this penalty.

To summarize, the proposed scheme consists of two steps. First, we perform a node pruning based on a PCA analysis, to keep the minimum number of nodes that generate practically the same information as the original ones. Then, we remove the remaining unimportant connections, as identified by those connections whose weight has an absolute value that is small compared with the rest of the connections of the same node. Retraining is applied only once after each of the two steps.

\section{Evaluation Methodology}\label{s:methodology}

Our goal is to prove that our pruning scheme, PCA+UC, provides pruning ratios that are similar or better than previous schemes in spite of not being iterative, unlike previous schemes, which makes them very costly or impractical. The schemes used for comparison are the \textbf{Baseline} where no pruning is applied, the \textbf{Near Zero Weights} using the connection pruning method described in Section~\ref{s:near_zero_pruning}, the \textbf{Input Weights Norm} which applies the node pruning method described in Section~\ref{s:avgw_pruning}, the \textbf{Similarity} node pruning method described in Section~\ref{s:sim_pruning}, and the \textbf{Random Weights} and \textbf{Random Nodes} which are simplistic methods that randomly chooses connections or nodes to prune given a target percentage. We do not include results for Scalpel (Section~\ref{s:scalpel_pruning}) and PCA Pruning (Section~\ref{s:pca_pruning}) since they perform worse than the above schemes.

To evaluate the pruning ratios achieved by PCA+UC we evaluate it on three state-of-the-art DNNs for two different application domains, one for acoustic scoring in speech recognition and two for image classification. We use the DNNs shown in Table~\ref{t:dnns}. The DNN for acoustic scoring is the one included in the Kaldi~\cite{Kaldi} toolkit, a popular framework for speech recognition. We have ported it to Tensorflow~\cite{Tensorflow} and trained it with the Librispeech~\cite{panayotov2015librispeech} dataset that contains 100 hours of speech. To evaluate \textit{LeNet5}, we use images of digits from the \textit{MNIST}~\cite{MNIST} dataset. In addition, we employ \textit{AlexNet}~\cite{AlexNetV1} in its second version~\cite{AlexNetV2} trained with the \textit{ImageNet}~\cite{ImageNet} dataset. All the networks and pruned models have been implemented in Tensorflow. For all the DNNs, we employ the whole test or validation dataset to obtain the accuracy.

\section{Results}\label{s:results}

This section evaluates the proposed PCA+UC pruning scheme, described in Section~\ref{s:method}, on different DNNs in terms of accuracy and amount of pruning. First, we compare our method with some previously proposed pruning strategies. Then, we present a sensitivity analysis of our scheme. 

The main benefit of PCA+UC is the time required to complete the pruning. For instance, AlexNet takes around 3 days to finish a retraining step on a GTX 1080 GPU. PCA+UC requires two retraining steps so it will take 2x the retraining time, i.e. less than a week for AlexNet. On the other hand, for the methods that require exploring multiple configurations per layer such as the Near-Zero pruning, finding the appropriate pruning percentages for all layers requires $n^l$ retraining steps, being $n$ the number of configurations analyzed per layer and $l$ the number of layers. Since AlexNet has eight layers, even for a very low value of $n$ such as 3, the pruning process would take around 54 years in a high-end GPU, or half a year in a farm of 100 GPUs. Considering that current DNNs, such as Resnet152 or Densenet201, include hundreds of layers the applicability of previous methods may not be feasible.

Table~\ref{t:kaldi_best_cmp} shows the pruning effectiveness for the Kaldi DNN. For each pruning scheme we report the maximum pruning we could achieve with negligible accuracy loss (less than 0.25\% in all cases). We can see that our method achieves the highest degree of pruning, resulting in a 70\% reduction of the weights and 70\% reduction of the number of computations. The next best scheme is the near-zero pruning, which achieves a 60\% reduction in both weights and computations. Since in Kaldi all the layers are fully-connected, the reduction in weights and the reduction in computations is the same. Note that our PCA+UC scheme only requires to retrain the DNN twice whereas for the other methods we have to carry out an iterative search to find the maximum percentage of pruning with negligible accuracy loss, and the DNN has to be retrained for each pruning percentage.

\begin{table}[t!]
\caption{Accuracy (WER) and percentage of weights and computations removed by different pruning schemes for the Kaldi DNN.}
\begin{center}
\begin{scriptsize}
\begin{sc}
\begin{tabular}{cccc}
\hline
 \multirow{ 2}{*}{\textbf{Pruning Method}} & \multirow{ 2}{*}{\textbf{WER(\%)}} & \textbf{Weights} & \textbf{FLOPS} \\
 & & \textbf{Pruned(\%)} & \textbf{Removed(\%)} \\
\hline
 Baseline & 10.04 & 0 & 0 \\
 Near Zero Weights & 10.18 & 60 & 60 \\
 Random Weights & 10.27 & 40 & 40 \\
 Input Weights Norm & 10.24 & 40 & 40 \\
 Similarity & 10.15 & 40 & 40 \\
 Random Nodes & 10.15 & 40 & 40 \\
 PCA+UC & 10.28 & 70 & 70 \\
\hline
\end{tabular}
\end{sc}
\end{scriptsize}
\end{center}
\vskip -0.25in
\label{t:kaldi_best_cmp}
\end{table}

Table~\ref{t:mnist_best_cmp} shows the results for the LeNet5 DNN. In this case, accuracy is measured as the top-1 so higher is better. Unlike Kaldi, in LeNet5 there are both convolutional and fully-connected layers. Most computations come from the convolutional layers while most of the weights are due to the fully-connected layers. Since some node pruning methods such as the similarity pruning can only be applied to fully-connected layers, they achieve a significant reduction in weights, but quite moderate in computations. We can observe that PCA+UC prunes 79\% of the weights and 52\% of the computations, which is the second best in terms of weight and computation reduction, only slightly below the near-zero.

\begin{table}[t!]
\caption{Accuracy (Top-1) and percentage of weights and computations removed by different pruning schemes for LeNet5.}
\begin{center}
\begin{scriptsize}
\begin{sc}
\begin{tabular}{cccc}
\hline
 \multirow{ 2}{*}{\textbf{Pruning Method}} & \multirow{ 2}{*}{\textbf{Top-1(\%)}} & \textbf{Weights} & \textbf{FLOPS} \\
 & & \textbf{Pruned(\%)} & \textbf{Removed(\%)} \\
\hline
 Baseline & 99.34 & 0 & 0 \\
 Near Zero & 99.33 & 80 & 80 \\
 Random Links & 99.31 & 60 & 60 \\
 Input Weights Norm & 99.37 & 87 & 20 \\
 Similarity & 99.4 & 87 & 20 \\
 Random Nodes & 99.29 & 87 & 20 \\
 PCA+UC & 99.41 & 79 & 52 \\
\hline
\end{tabular}
\end{sc}
\end{scriptsize}
\end{center}
\vskip -0.25in
\label{t:mnist_best_cmp}
\end{table}

Table~\ref{t:alexnet_best_cmp} shows the results for AlexNet. We can observe that PCA+UC prunes 67\% of the weights and 51\% of the computations. It does not prune the convolutionals as much as the previous heuristics so the reduction in computations is lower, but it achieves higher accuracy.

\begin{table}[t!]
\caption{Accuracy (Top-1) and percentage of weights and computations removed by different pruning schemes for AlexNet.}
\begin{center}
\begin{scriptsize}
\begin{sc}
\begin{tabular}{cccc}
\hline
 \multirow{ 2}{*}{\textbf{Pruning Method}} & \multirow{ 2}{*}{\textbf{Top-1 (\%)}} & \textbf{Weights} & \textbf{FLOPS} \\
 & & \textbf{Pruned(\%)} & \textbf{Removed(\%)} \\
\hline
 Baseline & 57.48 & 0 & 0 \\
 Near Zero & 56.55 & 80 & 87 \\
 Random Links & 56.46 & 50 & 63 \\
 PCA+UC & 59.7 & 67 & 51 \\
\hline
\end{tabular}
\end{sc}
\end{scriptsize}
\end{center}
\vskip -0.25in
\label{t:alexnet_best_cmp}
\end{table}

We have also evaluated our Unimportant Connections (UC) pruning alone to demonstrate that it is more effective than Near-Zero pruning. Table~\ref{t:cmp_link_pruning} shows the results for 70-90\% of pruning of the Kaldi DNN. We can see that UC achieves better accuracy with the same amount of pruning and, hence, taking into account the importance of the connections relative to each node is more effective than just considering the magnitude of all the weights on each layer.

\begin{table}[t!]
\caption{Link Pruning Comparison for the Kaldi DNN. (Baseline WER=10.04\%)}
\begin{center}
\begin{scriptsize}
\begin{sc}
\begin{tabular}{cccc}
\hline
\textbf{Weights Pruned (\%)} & \textbf{Near-Zero WER (\%)} & \textbf{UC WER (\%)} \\
\hline
 70 & 10.67 & 10.53 \\
 80 & 11.39 & 11.10 \\
 90 & 14.62 & 13.56 \\
\hline
\end{tabular}
\end{sc}
\end{scriptsize}
\end{center}
\vskip -0.25in
\label{t:cmp_link_pruning}
\end{table}

\subsection{Sensitivity Analysis}

The proposed pruning scheme uses two thresholds to control the amount of pruning and the loss of accuracy. Our key goal is to find values for these parameters that achieve high efficiency, i.e. large amount of pruning with negligible accuracy loss, for a wide range of DNNs. Therefore, the user does not have to manually tune these thresholds for each specific DNN, as it happens with other pruning schemes.

In our scheme, the first step (node pruning) is applied by keeping only as many neurons as PCA components are needed to represent 95\% of the original information (coefficient of variance). In the second step, connections are pruned based on its relative importance compared to the other connections of the same neuron.

We performed a sensitivity analysis using the Kaldi DNN since the accuracy of Kaldi is highly sensitive to model changes and the model size is reasonable. The parameters obtained from this analysis are used for all the networks, achieving the results shown in section~\ref{s:results}, which confirms that these parameters work well for different networks.

Table~\ref{t:kaldi_pca_variance} shows the results for the Kaldi DNN after applying the first step of our pruning method using different thresholds for the coefficient of variance (CV). The accuracy of Kaldi is measured as Word Error Rate (WER), so lower is better. As it can be seen, with 99\% of variance we can maintain the same accuracy or even slightly better, but the percentage of pruning is dramatically reduced to only 15\%. In contrast, if we use 90\% of variance, the pruning is quite high (64\%) but the accuracy is slightly affected (0.58\% loss). Finally, if we use 95\% of variance the accuracy is almost the same (only 0.2\% of loss) and the percentage of pruning is 45\%. Therefore, we use 95\% of coefficient of variance as the by default value for our scheme. In case that a small loss of accuracy could be assumed, 90\% of CV could be a good choice in order to achieve a higher degree of pruning.

\begin{table}[t!]
\caption{Kaldi results after the PCA step (first step) of the proposed pruning scheme using different coefficients of variance. (Baseline WER=10.04\%)}
\begin{center}
\begin{scriptsize}
\begin{sc}
\begin{tabular}{cccc}
\hline
 \multirow{ 2}{*}{\textbf{CV (\%)}} & \multirow{ 2}{*}{\textbf{WER (\%)}} & \textbf{Weights} & \textbf{FLOPS} \\
 & & \textbf{Pruned(\%)} & \textbf{Removed(\%)} \\
\hline
 99 & 10.00 & 15 & 15 \\
 95 & 10.24 & 45 & 45 \\
 90 & 10.62 & 64 & 64 \\
\hline
\end{tabular}
\end{sc}
\end{scriptsize}
\end{center}
\vskip -0.25in
\label{t:kaldi_pca_variance}
\end{table}

Table~\ref{t:kaldi_mean} shows the results after the second step of our pruning method using different thresholds for the percentage of the mean (computed as described in Section~\ref{s:pca_link_pruning}), which determines when a connection is pruned for the Kaldi DNN. We can conclude that 75\% is the most adequate threshold. A higher threshold can prune more links but looses some accuracy, whereas a lower threshold has about the same accuracy but is less effective at pruning.

\begin{table}[t!]
\caption{Kaldi results after the unimportant connections pruning step (second step) of the proposed pruning scheme for different thresholds of the mean. (Baseline WER=10.04\%)}
\begin{center}
\begin{scriptsize}
\begin{sc}
\begin{tabular}{ccccc}
\hline
 \multirow{ 2}{*}{\textbf{CV(\%)}} & \multirow{ 2}{*}{\textbf{Mean(\%)}} & \multirow{ 2}{*}{\textbf{WER(\%)}} & \textbf{Weights} & \textbf{FLOPS} \\
 & & & \textbf{Pruned(\%)} & \textbf{Removed(\%)} \\
\hline
 95 & 100 & 10.62 & 77 & 77 \\
 95 & 75 & 10.28 & 70 & 70 \\
 95 & 50 & 10.26 & 63 & 63 \\
\hline
\end{tabular}
\end{sc}
\end{scriptsize}
\end{center}
\vskip -0.25in
\label{t:kaldi_mean}
\end{table}

In summary, we set the by default pruning configuration to work with 95\% for the coefficient of variance to remove nodes and  75\% of the mean of the weights of each node to remove connections. Using this configuration PCA+UC achieves pruning percentages of 70\% in Kaldi, 79\% in LeNet5 and 67\% in AlexNet with negligible accuracy loss.

\section{Related Work}\label{s:related_work}

DNNs have become very popular in a wide range of environments and devices, from large data centers and high performance computers~\cite{chen2014dadiannao,XiongDHSSSYZ16a} to mobile devices~\cite{mobilecop,ShiDianNao}. DNNs are computationally and memory intensive, and consume a significant amount of energy. Therefore, custom architectures with optimizations such as pruning can provide important benefits. Some of the most popular applications of MLPs are speech recognition and machine translation~\cite{dahl2013improving,zhang2015deep}, whereas CNNs are commonly used for image classification~\cite{alwani2016fused}.

\textbf{DNN Optimizations}. Proposals for reducing the memory footprint and/or computations of DNNs include clustering~\cite{K-Means}, linear quantization~\cite{TPU} and pruning. Clustering uses methods such as K-means to reduce the number of different weights to K centroids. Each weight is then substituted by an index that corresponds to the closest centroid. Since the weights tend to be very similar, the number of centroids per layer can be kept relatively low (in the order of 16-256), which significantly reduces the storage requirements and memory bandwidth for the weights. However, computations still have to be performed in floating point by using the centroids and the total amount of computations is not reduced at all. Linear quantization maps each value, either weights or inputs, to a discrete set of values distributed over the whole range of possible values. Values are replaced by indexes, which identify the discrete set of values, and reduces their storage requirements. Unlike clustering, since the quantization is linear, most computations can be done by using directly the integer indices rather than the corresponding floating point values. The amount of computations is not reduced, but most computations are simpler since they operate on integer numbers rather than floating point. Therefore, quantization is particularly efficient in DNN accelerators to not only reduce the model size but also the energy consumption of most computations.

\textbf{Pruning}. As described in section~\ref{s:background} there is a large number of methods to prune DNNs. Some methods remove connections depending on the weights' values~\cite{Han2015}, others remove nodes taking into account the weights of each node~\cite{NodePruning,SimPruning}. PCA-based methods for fast pruning have been previously proposed~\cite{PruningPCA1, PruningPCA2, PruningPCA3, PruningPCA4, PruningPCA5, PruningPCA6, PruningPCA7}. However, they just project the weights and inputs using the principal components and do not apply retraining. The amount of pruning achieved with these PCA-based methods is extremely limited for modern DNNs. Most of these pruning methods are applied after training a baseline non-pruned network. On the other hand, some methods perform pruning during the training phase by learning the connections or nodes that are redundant~\cite{Scalpel,DynamicNetworkSurgery,EnergyAwarePruning}. We show in Section~\ref{s:pruning} that most of these previous methods are impractical due to the large number of parameters that have to be tuned for each DNN. Our proposal is inspired in previously proposed PCA-based techniques and weight pruning schemes and we show that it is highly effective and, more importantly, practical for a wide range of DNNs.

\section{Conclusions}\label{s:conclusions}

In this paper, we show that current DNN pruning schemes require that multiple parameters be configured by a trial and error process, often resulting in an exponential number of configurations to be evaluated, which may be impractical since each experiment requires a retraining of the network, which is an extremely costly operation. We propose the PCA+UC pruning scheme that overcomes this weakness. PCA+UC consists of two steps, one to remove redundancy at the neuron level, and another to remove redundancy at the connection level. The first step is based on a Principal Component Analysis (PCA) to remove nodes while the second step removes unimportant connections (UC) based on a novel metric that measures the importance of each link. The proposed scheme requires only two retraining operations and achieves results similar or even better than state-of-the-art iterative pruning methods.



\bibliographystyle{IEEEtran}
\bibliography{ref}

\end{document}